# OFFLINE HANDWRITTEN SIGNATURE IDENTIFICATION USING ADAPTIVE WINDOW POSITIONING TECHNIQUES

Ghazali Sulong[1], Anwar Yahy Ebrahim[2] and Muhammad Jehanzeb[3]

[1, 2, 3] UTM-IRDA Digital media centre (MaGIC-X), Faculty of Computing, Universiti Tecknologi Malaysia (UTM) Skudai 81310, Johor, Malaysia

[1] `ghazali@utmspace.edu.my`
[2] `anwaralawady@gmail.com`
[3] `mjsheikh2@live.utm.my`

*ABSTRACT*

*The paper presents to address this challenge, we have proposed the use of Adaptive Window Positioning technique which focuses on not just the meaning of the handwritten signature but also on the individuality of the writer. This innovative technique divides the handwritten signature into 13 small windows of size nxn (13x13). This size should be large enough to contain ample information about the style of the author and small enough to ensure a good identification performance. The process was tested with a GPDS dataset containing 4870 signature samples from 90 different writers by comparing the robust features of the test signature with that of the user's signature using an appropriate classifier. Experimental results reveal that adaptive window positioning technique proved to be the efficient and reliable method for accurate signature feature extraction for the identification of offline handwritten signatures .The contribution of this technique can be used to detect signatures signed under emotional duress.*

## 1. INTRODUCTION

The signature of any person is an important biometric characteristic which is usually implemented for personnel identification or document authentication. An increasing number of financial and business transactions are approved via signatures [1]. Handwritten signature as a method of authentication has become a part of everyday life, and there is probably not a single area where it is not used. Handwritten signature usage dates from ancient times and is held until today as a means of giving consent to something that needs to be done. The problem arises when someone decides to imitate the signature of the person with the purpose of fraud or false representation. Therefore, there is need for adequate protection personal signatures. One good method of protection is by the use of biometric systems based on handwritten signature [2].
Handwritten signature is defined as the first and last name written in your own handwriting [3]. It is often the case that the signature does not contain the full name but only one part of it, or sometimes a set of connected lines that do not resemble the name of the signer. This type of signature is called paraph, and is defined as an abbreviated signature, sometimes only the initial letter of the name, and placed on administrative act within the routine procedure, which means that whoever put the initials agrees to endorse the content. Work on signature can be carried out by using either signature identification or signature verification. Signature Identification can be done using a person's identity based only on biometric measurements. Signature verification is the process used to recognize an individual's handwritten signature for genuine or duplicate/forgery [4]. Signature identification and verification system are broadly classified in two ways: on-line and off-line. In an off-line technique, signature is signed on a piece of paper and scanned into a computer system. In an on-line technique, signature is signed on a digitizer and dynamic information like speed, and pressure is captured in addition to a static image of the signature. Verification decision is usually based on local or global features extracted from the signature being processed. Excellent verification results can be achieved by comparing the robust features of the test signature with that of the user's signature using an appropriate classifier [5].Due to the relative ease of use of an offline system, a number of applications worldwide prefer to use this system (e.g. Check verification in the bank) [6]. Even though the dynamic signature verification is more reliable as compared to offline systems in terms of accuracy, this online method requires special hardware digitizers, and pressure sensitivity tablet to capture the dynamic features, which the offline method does not require [7]. For an offline method, the extracted features of the individual's signature consider as an input to the system, which is then processed based on some predefined standardized methods. After the standardization stage, called pre-processing, extracted features from the offline handwritten signature are then identified and verified against a stored data set through a process known as graph matching. However, researchers in this field have failed to take into account external factors that may influence the signature of an individual as the individual may be under different emotions when signing his signature [8] [9] [10]. Thus, an individual though having a unique signature may sign a signature that can be in different shapes under different situations. Thus, it is important to take into account external factors when investigating a signature verification technique. This paper therefore attempts to address this challenge for offline handwritten signature identification by proposing a new technique based adaptive window positioning method for signature feature extraction that will give reliable accuracy in verifying an individual's signature even when the user is under different emotions since signature verification applications are used in our daily lives and will be exposed to human emotions. This proposed technique improves the efficiency and accuracy of offline handwritten signature verification and identification system. It also has a more enriched process of signature feature extraction by using the adaptive window positioning method. This gives it the ability to verify and identify an individual's signature even when signed under emotional stress. Furthermore, this study creates room for further research into the application of this technique in offline handwritten signature verification. The paper is organized as follows: Section 2 presents a review of related work to the study. Section 3 describes our proposed method. Section 4 discusses the results of our experiments using GPDS dataset. Section 5 concludes the paper.

## 2. BACKGROUND OF THE STUDY

Broadly, features can be classified as global or local, where global features represent signature's properties as a whole and local ones correspond to properties specific to a sampling point [11]. There are several feature extraction techniques employed in signature verification and identification systems [12] [13]. The basic purpose of any feature extraction is to extract accurate features from any given sample. The amount of feature extracted is irrelevant to the overall accuracy of the result as fewer features may be extracted using any techniques, and yet yield better result than one in which a large number of features are extracted. Thus the focus locus is the technique employed. Global and local features contain information, which are effective for signature recognition. Selection of different features is vital for any pattern recognition and classification technique [14]. Global signature features are extracted from the whole signature image. On the other hand, local geometric features are extracted from signature grids. Moreover, each grid can be used to extract the same ranges of global features. The Combination of these global and local features is further used to successfully determine the identity of authentic and forged signatures from a database. This set of geometric features is
Used as input to the identification system.

Table 1. List Different Offline Handwritten Signature Identification and Verification Methods

| S/N | Approach | Characteristics | Advantages | Disadvantages |
|---|---|---|---|---|
| 1. | Template Matching approach | - Employs pattern comparison process | - Suitable for detecting genuine signatures via rigid matching | - Not appropriate for detecting skilled forgeries |
| 2. | Neural Networks (NN) approach | - Learns by example thus good for learning the underlying structure of the data set.<br>- Can be used to model complex functions<br>- Highly suitable for modeling global features of handwritten signatures | - Widely accepted classifiers for pattern recognition problems<br>- Has very low FAR and FRR results | - Not very suitable for modeling statistical and geometric features<br>- Requires a highly representative data set |
| 3. | Hidden Markov Models (HMM) approach | - Best suited for sequence analysis in signature verification<br>- Uses stochastic matching (model and signature) to extract variability between patterns and their similarities<br>- Has various topologies and adopts probability density function modeling in its design for the verification task | - Can easily detect simple and random forgeries in signature verification | - Very poor in detecting skilled forgery |
| 4. | Statistical approach | - Employs statistical method to determine the relationship, deviation, etc between two or more data items<br>- Uses the concept of Correlation Coefficients | - Good at identifying random and simple forgeries<br>Its :graphometry -based approach avails so many usable features for signature verification, e.g., calibration, proportion, guideline and base behavior | - Its use of static features limits it from detecting skilled forgery |
| 5. | Structural and Syntactic approach | - Uses symbolic data (e.g. Signatures) structures such as strings, graphs, and trees to represent recognition patterns<br>- Employs the use of a Modified Direction Feature (MDF) to extract transition locations | - Appropriate for detecting genuine signatures and targeted forged signatures | - Very exhaustive method as it requires large computational efforts and training sets |
| 6. | Wavelet-based approach | - It is a multi-resolution transform that can decompose a signal into lowpass and highpass information<br>- Wavelet theory is employed in decomposing a curvature-based signature into a multi-resolution signal | - Can be applied in both offline and online signature verification<br>- Can decompose a curvature-based signature into a multi-resolution format<br>- Can be applied to symbolic languages such as Chinese and Japanese besides English | |

The global features that are extracted from the signature sample include [13]: Width (Length), Height, Aspect ratio [15], Horizontal projection, Vertical projection, Area of black pixels, Normalized area, Center of gravity, Maximum and minimum black pixels in vertical projection, Maximum and minimum black pixels in horizontal projection, Global baseline, Upper and lower edge limits, Middle zone, Hough transform [16], Curvelet transform. Local features- also known as Grid features can be extracted from gray level, binary and thinned signature images. From the small regions of the whole image, local features are estimated, such as center of gravity, width, height, horizontal and vertical projections, aspect ratio, area of black pixels of each grid region, normalized area of black pixels, gradient and concavity features etc . The global features can also be considered as local features for each grid region. To obtain a set of global and local features, both of these feature sets are combined into a feature vector and the feature vector is sent as input to the classifiers for generating matching scores [16] [2]. Other offline signature features include Statistical features [10], Geometry and topological features [17], Kurtosis [18], Orientation [18], Gabor Wavelet [19], Eccentricity [18], Skewness [15], Discrete Wavelet Transform (DWT) Features [20], Modified Direction Feature [15], and Contour let transform (CT) [21] [15].

## 3. PROPOSED METHOD

In this section, we present our method and its application processes to offline signature feature extraction using the adaptive window positioning technique. Thus, the section illustrates the proposed methodology for the offline handwritten signature identification system using adaptive widow positioning technique. Here, we follow through all the logical steps involved starting from getting the handwritten offline signature image to the signature identification stages. Figure 1 presents a flowchart of our proposed system.

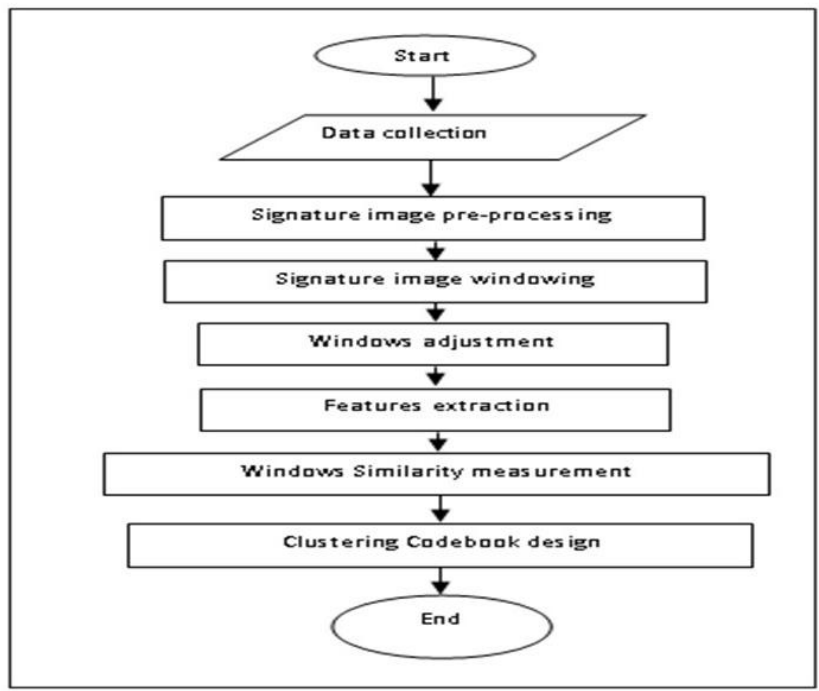

Figure 1. Flowchart of offline signature identification process using window positioning technique

## 3.1. Data Acquisition and Pre Processing

This consists of the data collection and signature image pre-processing stages. Our data was collected from GPDS dataset because it is one of the well-known and widely used databases in signature identification and verification applications and researches. GPDS dataset is a standardized database of offline English signature images containing 4870 signature samples of 90 different writers. The pre-processing stage is responsible for improving the image quality, and precedes the signature fragmentation phase. It makes extraction and identification of the individual features much easier by improving the identification performance. Sometimes, some of the original information is lost during the clean-up process of the background noise of the digitized image. For the purpose of our study, a global threshold of Otsu's algorithm was used in this study to convert the selected dataset images into binary images of the signature. This binary image serves as input to the next step.

## 3.2. Signature Image Windowing

In order to extract the writing patterns used frequently by an individual, we first implement division (segmentation) on the signature to produce small sub-images (fragments). This division is carried out by positioning small windows over the signature in a way that will exploit most of the redundancy in the signature and produce signature fragments that allows for meaningful comparison of these fragments. We have chosen to carry out this division in square windows of nxn; this size should be large enough to contain ample information about the style of the author and small enough to ensure a good identification performance. In this paper, the window size that has been used is fixed at a value of 13 because a window size (13x13) can be applied to any input image to get the optimum output. This adaptive window positioning method seeks to follow the ink trace with the objective of achieving an optimal window positioning that is based on the analysis of the skeleton of the handwritten signature image with respect to the drawing. This is made possible by removing the pixels on the boundaries of the component without breaking it apart, then placing the first window on the original component by finding the first image pixel (the starting point of the written component) and scanning the pixel from top to bottom and left to right as shown in Figure 2a. The second step is to copy the same placed window on the skeleton image and define a four-flag direction (East, West, North and South) for each window, and set the related flag if the skeleton exits from that particular side as shown in Figure 2b.

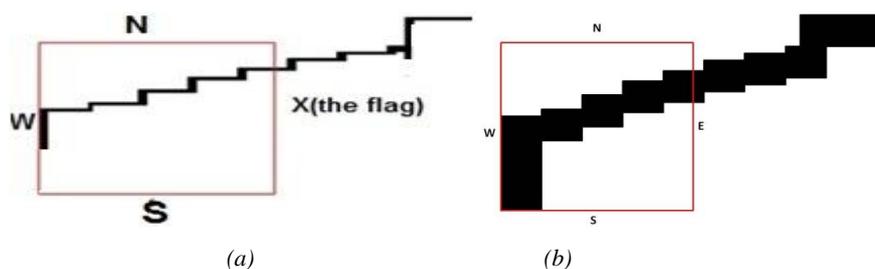

*(a)*  *(b)*
Figure 2. (a) Skeleton trace. (b) Window positioning on a component

If the skeleton exits from the E, then the next window must be placed towards the right of the current window (on the original component) as shown in Figure 3a, and then shift the window in a vertical direction (up and down) to find its best position with respect to the text trace as in Figure 3b, which shows how the window is placed to the right of an existing window and needs to be moved in the vertical direction to find its final position. On the other hand, if the skeleton exists from the N, then the next window must be placed on top of the current window and moved horizontally (left or right) so as to be well placed over the text, In some cases where the skeleton exists from more than one side, then the dividing process will treat each of the branches separately.

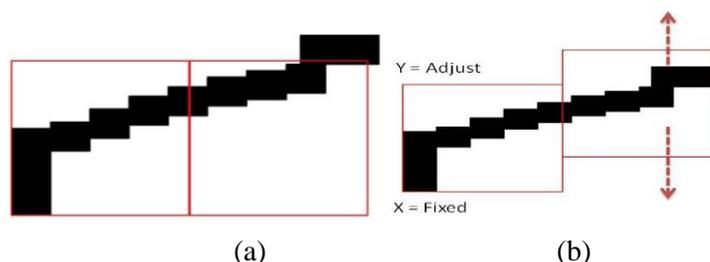

(a) (b)

Figure 3. (a) Initial position of the next window. (b) Window sliding with respect to the text trace (well placed window).

## 3.3. Windows Adjustment

Before comparing the similarities between the sub-images (fragments) and saving it on clusters, the patterns must first be adjusted inside its window in order to prevent the disadvantages of direct comparison which keep nxn pixel values (for a window size of n) thereby making the comparisons susceptible to noise and distortion. Thus, an adjustment is done by repositioning the image trace with respect to its window. We move the shape towards the upper-left corner of the window as shown in Figure 4, such that only features that are independent of the window positioning style are computed (in direct comparison). It is important to note that this window adjustment does not include pattern rotation because the rotated versions are not considered since they are not produced by the same gesture of the hand and thus should not be used or grouped in the same class. Also we took into consideration the fact that same scale exists for handwritten signature images within same sample, since the same writer cannot change the writing scale on the data set. Thus, two sub-images cannot be compared using this feature because they are not necessarily scale invariant due to the dataset standardization. We therefore assume that all images are almost of the same image size and scanned in the same resolution.

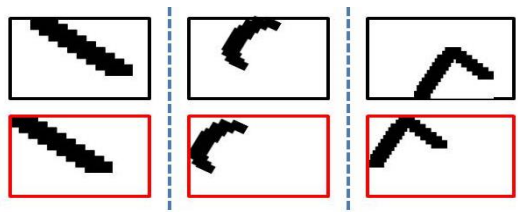

Figure 4. Pattern adjustments inside its window

## 3.4. Features Extraction

Here, we extract the set of features (shape measures) from the patterns and represent the images in a feature space. Since a typical data analysis problem involves many observations as well as a good number of respective features, it is important to organize such data in a sensible way before it can be presented and analysed by humans or machines. The aim of shape characterization is to obtain shape measures to be used as features for classification in patterns.
For our study, we used "adaptive window positioning" technique for features extraction. This technique, first proposed by (Siddiqi, Vincent, 2007), is more adapted to the ink trace and has the flexibility to occupy a line position within a window. Since the images are offline, it is not possible to follow the stroke trajectory that is followed by the writer. Nevertheless, this method to some extent seeks to follow the ink trace with the objective of achieving an optimal window positioning that is based on the analysis of the skeleton of the handwritten Signature image with respect to the drawing.

### 3.5 Similarity Measurement

After representing the sub-images from a set of features, then it needed to do a similar measure among all the windows which enables the comparison between two sub-images (fragments). The two sub-images are compared with the following correlation similarity measurement:

$$S(X,Y) = \frac{n_{11}n_{00} - n_{10}n_{01}}{[(n_{11}+n_{10})(n_{01}+n_{00})(n_{11}+n_{01})(n_{10}+n_{00})]^{1/2}} \quad (1)$$

Where, Nij is the number of pixels of the two sub-images X and Y, which they have the values i and j respectively, at the corresponding pixel positions. This measure will be close to 1 if the two compared sub-images are similar and in extreme case it will have a value equal to 1 indicating that the two shapes are exactly the same.

### 3.6 Clustering to get the Code book

On this step is the clustering of the adjusted patterns of the sub-images will be done. The objective is to group similar patterns in the same class, these classes would then correspond to the frequency patterns occurring in writing. The number of elements per class, however, depends upon the amount of text in the sample, so the sufficient number cannot be a fixed value. (See figure 7 for more details).

## 4. DISCUSSION OF RESULT

In this section, the initial results of the proposed method have been applied and illustrated to verify the validity of the earlier mentioned procedural methodology in section 3. The work is implemented using Delphi programming language under Microsoft Windows operating system environment and a demonstration of the performance is reported at the end of this section

### 4.1. Pre Processing Information

This is the first step carried out on the scanned signature image. In this step, pre-processing mechanisms describe converting the image from a grey level image into binary image with minimal consideration of the noise model [22]. From the image and to make it clearer and more useful for the signature identification process. Figure 5 shows the before and after pre-processing for a signature image.

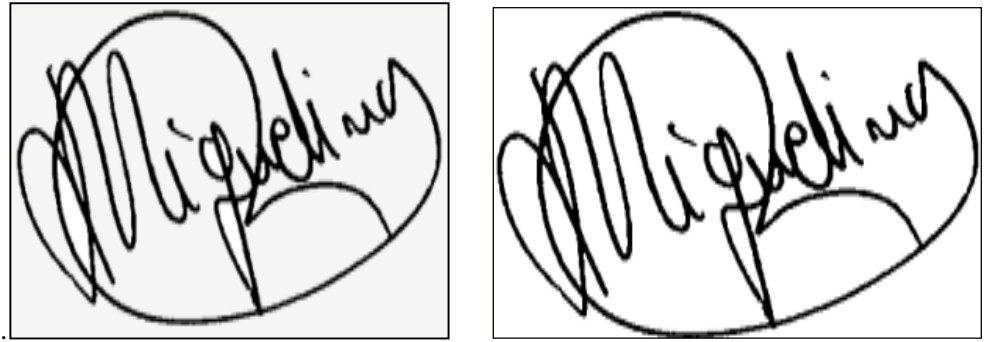

*(a)* *(b)*

Figure 5. (a) And (b) are before and after the pre-processing respectively

### 4.2. Windowing

After identifying the components in an offline signature image, a division (windowing) procedure was carried out on each component from top-bottom (vertical) and left-right (horizontal) onward directions of the image trajectory, divided by (13x13) window size (see Figure 6a). Figure 6b shows a magnified photo of the signature, which gives a clearer view of the windows division all around the image following the trace trajectory with a well-positioned window indicating that no overlap exists.

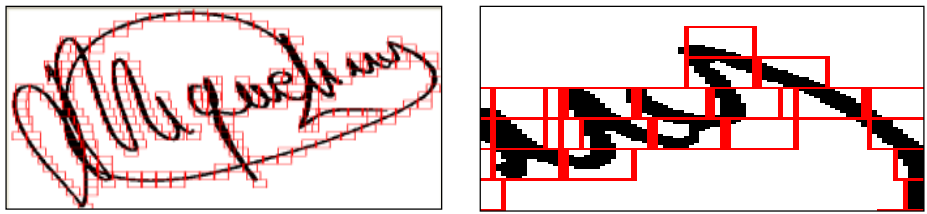

(*a*) *(b)*

Figure 6. (a) Offline signatures image division. (b) A closer look at the windows for the signature part

### 4.3. Pattern Adjustment

Before moving to the features extraction stage, we adjusted pattern parts inside each window by moving it to the upper left corner. This makes the calculation of the feature extraction processes more accurate and easier. Figure 7a shows the patterns extracted from a handwritten signature image of each window, while Figure 7b illustrates the patterns inside each window after applying the necessary adjustment on each pattern.

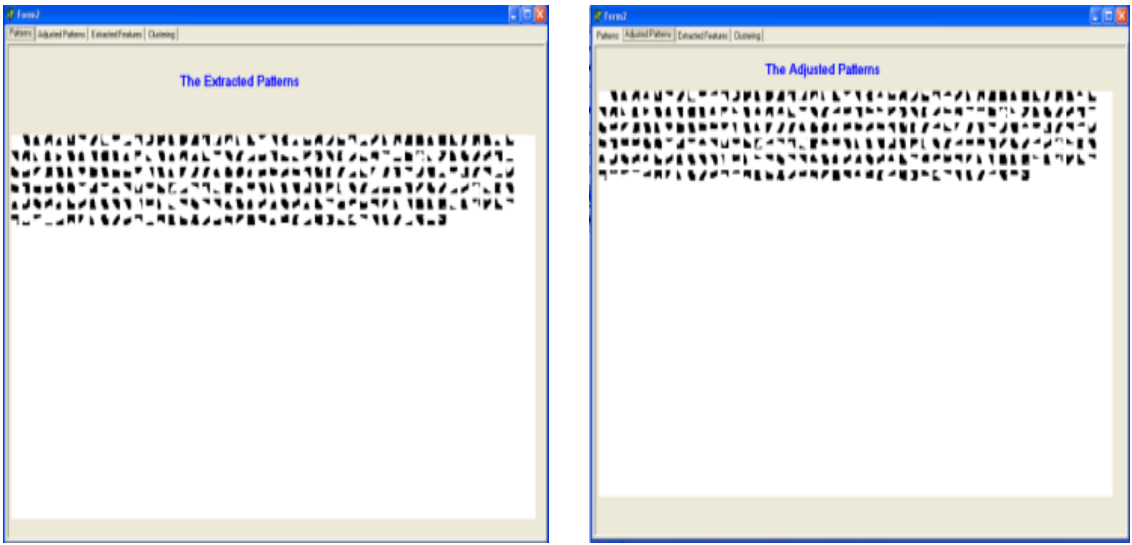

*(a)* *(b)*

Figure 7. (a) Shows the extracted patterns. (b) Shows the extracted patterns after making the adjustment

## 4.4. Features Extraction

This last three stages are considered the most important phases among all the procedural stages, due to their impact on the effectiveness of the identification and verification process, the more related features extraction leads to better categorized codebooks, as a consequence it results in a better identification process. Figure 8 shows the result of the feature extraction after applying the techniques mentioned in section 3. For instance, the row zero represents the first window, and under the HH3 column, is the value 10 which represents the frequency of pattern that accrued in the signature (please refer to section 3.6). The higher the value, the more it shows a specific pattern with the original signature in the data set, which implies that this is a high similarity between the test signature and the data set signature.

Figure 8. Shows the feature extraction

## 4.5. Similarity Measurement

Here, we classify the many features extracted from the input image into groups in terms of their similarity attributes. A similarity function mentioned earlier in section 3 was coded in the software to determine the range of values of the features variety and then classified into groups according to a specific threshold magnitude (Figure 8).

## 4.6. Clustering Codebook

Clustering was used to classify the extracted features into classes based on our similarity function. The number and length of classes varied depending on the authorship of the writer and the signatures differences. The result of our experiment is illustrated in Figure 9, which shows the clusters of a given input handwritten signature image, with a total number of 47 classes (c1 to c47).

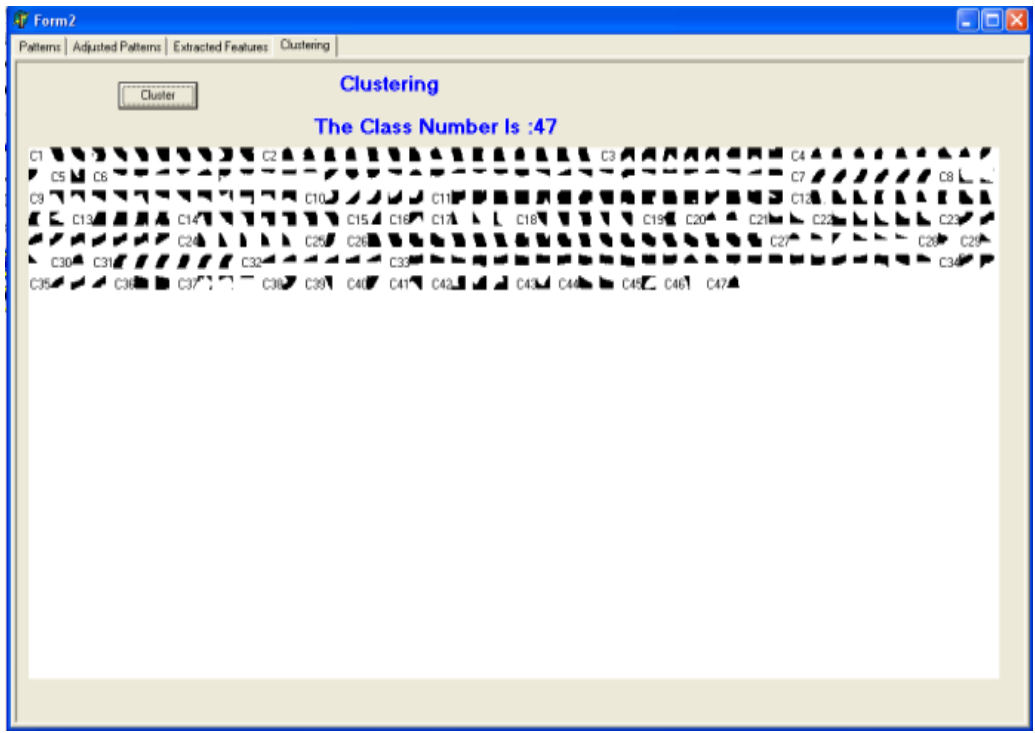

Figure 9. Sample Codebook for one Signature using our proposed Adaptive Window Positioning Technique (A cluster of 47 classes).

## 4.7. Validation of Result

The validation of our process was tested with a GPDS dataset containing 4870 signature samples from 90 different writers by comparing the robust features of the test signature with that of the user's signature using an appropriate classifier. GPDS is well known and one of the most widely used database for offline signature identification and verification. In order to evaluate the performance of the proposed method, we use three standard evaluation criteria: False Acceptance Rate (FAR), False Rejection Rate (FRR), and Equal Error Rate (ERR).

Table 2. Shows the test result comparison of different methods with our proposed method on 1200 signature of 40 users.

| Authors | Methods | FAR (%) | FRR (%) | EER (%) |
|---|---|---|---|---|
| Proposed method (2014) | Adaptive Window Positioning Techniques | 8.68 | 6.12 | 7.40 |
| Vargas et. al., 2011 | High Pressure Polar Distribution | 14.66 | 10.01 | 12.33 |
| Larkins et. al. 2009 | Adaptive Feature Thresholding | 10.96 | 8.16 | 9.66 |
| Nguyen et. al., 2009 | Global Features for Offline Systems | 17.25 | 17.26 | 17.25 |
| Chen et. al., 2006 | Graph Matching | 16.30 | 16.60 | 16.40 |

We used the FAR to account for all the skilled forgeries, FRR for only genuine signatures, and ERR for verification. The ERR is computed from the average of the values derived from both FAR and FRR and is a very good criterion for evaluating the accuracy of a method. The method with the lowest ERR can be considered as the most accurate technique. Hence, results of our experiment (Table II) reveal that adaptive window positioning technique proved to be the efficient and most reliable method for accurate signature feature extraction for the identification of offline handwritten signatures.

## 5. CONCLUSION

This paper proposed the use of the adaptive window positioning technique for offline handwritten signature identification. It employed the division of signature images into 13x13 windows and created some new cluster patterns for each window when classified into groups of similar attributes. The results of our study show that the adaptive window positioning technique is a more efficient and reliable method for accurate offline handwritten signature feature extraction. The major contribution of this paper is that this technique can be employed by signature identification and verification systems for the detection of offline handwritten signatures signed under duress or emotional stress.

## ACKNOWLEDGMENT

We wish to acknowledge the Faculty of Computing, Universiti Tecknologi Malaysia (UTM) for their support to this research.

# REFERENCES

[1]     B. Miroslav, K. Petra, F. Tomislav, (2011) "Basic on-line handwritten signature features for personal biometric authentication", MIPRO, Proceedings *of the 34th International Convention ,Opatija,* May 2011, pp. 1458-1463.

[2]     D. R. Shashikumar, K. B. Raja, R. K. Chhotaray, S. Pattanaik, (2010) "Biometric security system based on signature verification using neural networks", *Computational Intelligence and Computing Research* (ICCIC), (Bangalore),  pp 1-6.

[3]     G. A. Khuwaja and M. S. Laghari, (2011)"Offline   handwritten signature recognition*", World Academy of Science, Engineering and Technology*, vol. 59, pp. 1300-1303.

[4]     H. B. Kekre, V. A. Bharadi, (2010) "Gabor filter based     feature vector for dynamic signature recognition", *International Journal of Computer Application*, vol. 2 (3), May, pp 74-80.

[5]     Y. M. Al-Omari, S. H. S. Abdullah, and K.  Omar, (2011) "State-of-the-art in offline signature verification system", *International Conference on Pattern Analysis and Intelligent Robotics,* June 2011, Putrajaya, (Malaysia), pp. 59-64.

[6]     S. Arora, D. Bhattacharjee, M. Nasipuri , L. Malik , M. Kundu and D. K. Basu, (2010) "Performance comparison of SVM and ANN for handwritten Devnagari character recognition", *International Journal of Computer Science Issues*, vol. 7 (3), May, pp. 1-10.

[7]     E. Alattas, and S. Meshoul, (2011)"An effective feature selection method for on-line signature based authentication", *Eighth International Conference on Fuzzy Systems and Knowledge Discovery* (FSKD), (Shanghai), vol. 3, July, pp.1431-1436.

[8]    S. A. Daramola, T. S. Ibiyemi , (2010)"Offline signature recognition using Hidden Markov Model   (HMM)", *International Journal of Computer Applications*, vol. 10 (2), November, pp. 17-22.

[9]     S. K. Shrivastava, S.  Gharde, (2010) "Support vector machine for handwritten Devanagari numeral   recognition" *International Journal of Computer Applications* (0975 – 8887), vol. 7 (11), October.

[10]    D. Samuel, and I.Samuel , (2010) "Novel feature extraction technique for off-line signature verification system", *International  Journal of Engineering Science and Technology* ,vol. 2 (7), pp. 3137-3143.

[11]     S. K. Shrivastava, and S. S. God, (2010) "Support vector machine for handwritten Devanagari numeral recognition" *International Journal of Computer Applications* (0975 – 8887), vol. 7 (11), October.

[12]    B. Singh, A. Mittal, and D. Ghosh, (2011)"An evaluation of different feature extractors and classifiers for offline handwritten Devnagari character recognition", *Journal of Pattern Recognition    Research*, pp. 269-277.

[13]    D. R. Kisku, P. Gupta, and J. K. Sing, (2010)"Offline signature identification by fusion of multiple    classifiers using statistical learning theory", *International Journal of Security and its Applications*, vol. 4 (3), July, pp. 35-45.

[14]    Siddiqi and Vincent, (2010)"Text independent writer recognition using redundant writing patterns    with contour-based orientation and curvature features", *Pattern Recognition*, vol. 43, pp. 3853–3865.

[15]    Vu Nguyen, M. Blumenstein, V. Muthukkumarasamy, and G. Leedham, (2007)"Off line signature   verification using enhanced modified direction features in conjunction with neural classifiers and support vector machine", *Ninth International Conference on Document Analysis and Recognition* , Vol 2, pp. 734-738.

[16]    O. Mirzaei, H. Irani , H. R. Pourreza , (2011)"Offline signature recognition using modular neural networks with fuzzy response integration", *International Conference on Network and Electronics Engineering* IPCSIT, Vol. 11, (Singapore), pp. 53-59.

[17]     M. K. Kalera, S. Sriharly, A. Xu, (2004) "Offline signature verification and identification using distance    statistics", *International Journal of Pattern Recognition and Artificial Intelligence*, vol. 18 (7),  pp. 1339-1360.

[18] S. M. Odeh and M. Khalil, (2011) "Off-line signature verification and recognition: Neural Network    Approach", *International Conference On Innovations In Intelligent Systems And Applications* (INISTA), pp. 34-38.

[19] M. H. Sigari, M. R. Pourshahabi, and H. R. Prize, (2011)"Off-line handwritten signature identification  and verification using multi-resolution Gabor wavelet", *IJBB*, vol. 5, pp.1-15.

[20] M.S. Shirdhonkar and M. Corker, (2011)"Off-line handwritten signature identification using rotated  complex wavelet filters", *International Journal of Computer Science Issues*, Vol. 8 (1), January , pp. 478-482.

[21] M. R. Pourshahabi, M. H. Sigari, and H. R. Prize, (2009)"Offline handwritten signature identification  and verification using contourlet transform", *International Conference on Soft Computing and Pattern Recognition*, Malacca,(Malaysia), December,  pp. 670-673.

[22] M. Rama Bai, (2013)"Composite Texture Shape Classification Based on Morphological Skeleton and Regional Moments "*Signal & Image Processing: An International Journal* (SIPIJ), India, Vol.4, No.3, June, pp. 4313.

**AUTHORS**

**Ghazali Sulong** received his BSc degree in statistic from National University of Malaysia, In 1979, and MSc and PhD in computing from University of Wales , Cardiff , United Kingdom , in 1982 and 1989, respectively.  He  is currently a  professor at the  Faculty  of Computing , Universiti Teknologi  Malaysia . His research  interest  includes Biometric – Fingerprint Identification , face  recognition , iris verification, ear recognition, handwriting Recognition, and writer  identification ; object  recognition ; medical  image segmentation , Enhancement and restoration; human activities recognition; data hiding - digital watermarking And Steganography ; image encryption; image compression ; image fusion ; image mining ; Digital image forensics; object detection, segmentation and tracking.

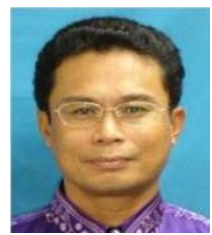

**Anwar Yahya Ebrahim**  received  her  B.Sc. degree from Babylon University, Iraq  in  2000. And her M.Sc. degree from MGM College Dr. Babasaheb Ambedkar Marathwada University, India in 2009, currently she is PHD.Student in Universiti Technology Malaysia (UTM).  Her  Research  interests  include signatures  identification and verification , features  recognition , Image analysis and classification.

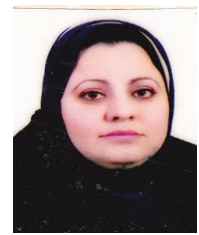

**Muhammad Jehanzeb**  received  his  B.Sc. degree from Arid Agriculture University , Pakistan in 2005, his M.S. degree from Iqra University, Pakistan, in 2007. He earned his Ph.D. at Universiti  Teknologi Malaysia (UTM) in 2013. Presently, he is working as Post-Doctoral Fellow at Universiti Teknologi Malaysia (UTM).  His research interests include document analysis and recognition, pattern recognition, image analysis and classification.

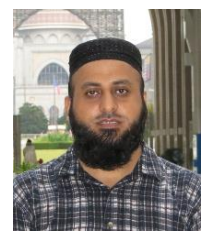